# PhaseNet for Video Frame Interpolation

Simone Meyer[1,2]    Abdelaziz Djelouah[2]    Brian McWilliams[2]    Alexander Sorkine-Hornung[2*]
Markus Gross[1,2]    Christopher Schroers[2]

[1]Department of Computer Science, ETH Zurich    [2]Disney Research

simone.meyer@inf.ethz.ch    aziz.djelouah@disneyresearch.com

## Abstract

*Most approaches for video frame interpolation require accurate dense correspondences to synthesize an in-between frame. Therefore, they do not perform well in challenging scenarios with e.g. lighting changes or motion blur. Recent deep learning approaches that rely on kernels to represent motion can only alleviate these problems to some extent. In those cases, methods that use a per-pixel phase-based motion representation have been shown to work well. However, they are only applicable for a limited amount of motion. We propose a new approach, **PhaseNet**, that is designed to robustly handle challenging scenarios while also coping with larger motion. Our approach consists of a neural network decoder that directly estimates the phase decomposition of the intermediate frame. We show that this is superior to the hand-crafted heuristics previously used in phase-based methods and also compares favorably to recent deep learning based approaches for video frame interpolation on challenging datasets.*

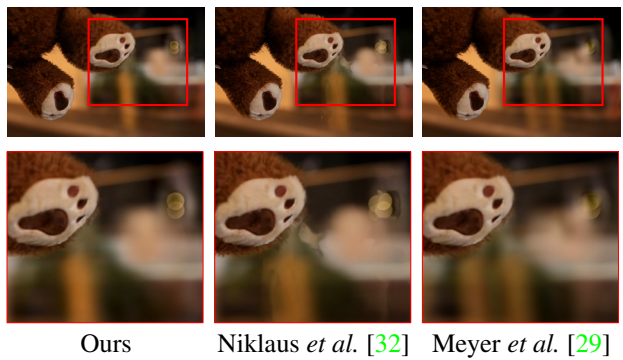

Figure 1: **Video frame interpolation.** Compared to recent kernel based method [32], our approach is able to handle complex scenarios containing motion blur or light changes. It also improves over existing phase-based interpolation methods [29] relying on heuristics, which are limited in their motion range. (Image source: [22])

## 1. Introduction

Video frame interpolation is a classic problem in video processing and has many applications ranging from frame rate conversion to slow motion effects. Traditionally this problem is formulated as finding correspondences between consecutive frames which are then used to synthesize the in-between frames through warping. These methods [5, 37, 41] usually suffer from the inherent ambiguities in estimating the correspondences and are particularly sensitive to occlusions/dis-occlusion and changes in color or lighting.

To overcome the limitations of traditional methods two main directions have been explored. The first [8, 29] relies on phased-based decomposition of the input images, but methods in this category are limited in the range of motion they can handle. The second direction is based on recent advances in deep learning [32]. These methods have largely improved over optical flow based methods, but are still not able to handle challenging scenes containing light changing and motion blur.

In this work we propose a novel neural network architecture, PhaseNet, which combines the phase-based approach with a learning framework. PhaseNet mirrors the hierarchical structure of the phase decomposition which it takes as input. It then predicts the phase and amplitude values of the in-between frame level by level. The final image is reconstructed from these predictions at different levels. Therefore, PhaseNet is able to handle a larger range of motion than existing phase-based methods [29] (which use hand-tuned parameters) while addressing the issues of optical flow and kernel based methods [32].

PhaseNet processes channels of the input images independently and shares weights across channels and pyramid levels and as such requires a relatively small number of parameters.

Furthermore, we introduce a *phase loss*, which is based on the phase difference between the prediction and the ground truth and encodes motion relevant information. To improve training efficiency and stability, PhaseNet is trained hierarchically starting from the coarsest scale and proceeding incrementally to the next finest scale. Altogether, we show that this allows us to outperform existing state-of-the-art methods for video frame interpolation in challenging scenarios.

---

[*]Alexander Sorkine-Hornung is now at Oculus. He contributed to this work during his time at Disney Research.

## 2. Related Work

Intermediate frames of a video sequence are commonly obtained by interpolating an optical flow field [5] representing a dense correspondence field between images. Therefore the final interpolation result is heavily dependent on the accuracy of the computed flow. However, finding a pixel-accurate mapping is an inherently ill-posed problem. Existing approaches usually require computationally expensive regularization and optimization, see [41] for a thorough analysis. Furthermore, they often rely on the brightness constancy constraint and therefore have difficulties handling scenes with large changes in brightness, although small changes can be handled by working in the gradient domain [26]. Alternatevly, Fleet *et al.* [11] suggest to use a phase constancy constraint to compute the optical flow and recently, a pure phase-based interpolation method was proposed [29]. By using per-pixel modifications and not computing explicit correspondences, such an approach is more stable to lighting changes. Its main drawback is the limit in the range of motion and the heuristics it introduces. Phase-based motion representations have also been used for various other applications, such as motion magnification [45, 10, 48], light-fields [49], image editing [28] and image animation [36]. Approaches to extend the motion range have been proposed, e.g., by combining it with optical flow [10] or by computing a disparity map [49]. In this work we increase the robustness by combining it with a neural network.

Neural networks have enjoyed a recent resurgence in popularity due to the huge growth in data and computational resources which has allowed models to be trained successfully [21, 6]. They have achieved state-of-the-art performance in a variety of applications domain such as large-scale image and video classification, detection, localization and recognition (e.g. [20, 40, 42, 15]). Most models for these tasks are trained in a supervised manner, requiring large amounts of labeled data. Supervised methods [9, 16, 43] have also been suggested for optical flow estimation. However, this requires a large volume of ground-truth optical flow data. To estimate optical flow without ground-truth data Long *et al.* [24] synthesize interpolated frames as an intermediate result.

Neural networks have been applied for image synthesis in various contexts [27, 12, 18]. Directly predicting images often produce blurry results [14, 44, 47]. Instead of pre-dicting pixel value, Zhou *et al.* [50] predict an *appearance flow* and use it to warp pixels and synthesize novel viewpoints. In the same spirit, Liu *et al.* [23] propose to train a convolution neural network to synthesize an intermediate frame by flowing and blending pixel values from the existing input frames according to the predicted voxel flow. Niklaus *et al.* [32, 31] combine motion estimation and image synthesis into a single convolution step. These methods generally result in sharp images and already better handle challenging situations—such as brightness changes—than traditional optical flow methods. However, in these scenarios, we show that our phase-based approach performs better.

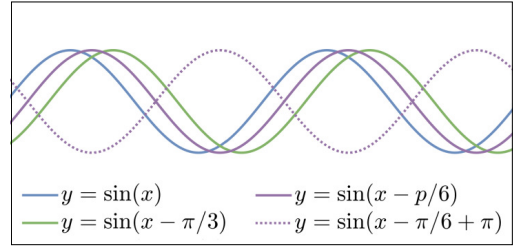

Figure 2: **Interpolation as phase shift.** The translation of a simple sinusoidal function (blue to green) can be expressed by the phase difference. To estimate the middle signal, phase-based interpolation needs to determine the correct phase value among the two possible solutions in purple.

## 3. Motion Representation

Similar to previous works, we base our method on the intuition that motion of certain signals can be represented by the change of their phase [29, 45]. Our goal is to directly estimate the phase value of the intermediate image. To illustrate our motivation, we adapt the example used in [29] followed by a similar review of the phase-based image decomposition for completeness.

**Motivation.** We first introduce the concept and challenges of phase-based motion representation. To illustrate them we use one dimensional sinusoidal functions $y = A\sin(\omega x - \phi)$, where $A$ is the amplitude, $\omega$ the angular frequency and $\phi$ the phase. Assuming we have two functions, which are defined as $y = sin(x)$ and $y = sin(x - \pi/3)$, for example. Graphically they represent the same sinusoidal function but one is translated by $\pi/3$, see Figure 2. The translation, i.e. the motion, can be represented by the phase difference of $\pi/3$. This demonstrates the general idea of representing motion as a phase difference. In terms of frame interpolation, these two curves (blue and green) would correspond to the input images. An in-between curve would then represent the interpolated intermediate image. But due to the $2\pi$-ambiguity of phase values (i.e. $y = sin(x - \pi/3) = sin(x - \pi/3 + 2\pi)$) there exists two valid solutions, namely $y = sin(x - \pi/6)$ (purple) and $y = sin(x - \pi/6 + \pi)$ (purple dotted). The difficulty of phase-based frame interpolation is to determine, which is the correct solution. While [29] describes a heuristic on how to correct the phase difference to correspond to the actual spatial motion. In this work we propose to learn to directly predict the phase value of the desired intermediate result.

**Image decomposition.** More complex one dimensional functions can be represented in the Fourier domain as a sum of complex sinusoids over all frequencies $\omega$:

$$f(x) = \sum_{\omega=-\infty}^{\omega=+\infty} A_\omega e^{i\phi_\omega} \ . \quad (1)$$

Images can be seen as two dimensional functions which can be represented in the Fourier domain as a sum of sinusoids over not only different frequencies but also over different spatial orientations. This decomposition of the image can be obtained by using e.g. the complex-valued steerable pyramid [35, 38, 39]. By applying the steerable pyramid filters $\Psi_{\omega,\theta}$, consisting of quadrature pairs, we can decompose an image into a set of scale and orientation depended complex-valued subbands $R_{\omega,\theta}(x,y)$:

$$R_{\omega,\theta}(x,y) = (I * \Psi_{\omega,\theta})(x,y) \quad (2)$$
$$= C_{\omega,\theta}(x,y) + i\, S_{\omega,\theta}(x,y) \quad (3)$$
$$= A_{\omega,\theta}(x,y)\, e^{i\phi_{\omega,\theta}(x,y)} \ , \quad (4)$$

where $C_{\omega,\theta}(x,y)$ is the cosine part and $S_{\omega,\theta}(x,y)$ the sine part. Because they represent the even-symmetric and odd-symmetric filter response, respectively, it is possible to compute for each subband the amplitude

$$A_{\omega,\theta}(x,y) = |R_{\omega,\theta}(x,y)| \quad (5)$$

and the phase values

$$\phi_{\omega,\theta}(x,y) = \text{Im}(\log(R_{\omega,\theta}(x,y))) \ , \quad (6)$$

where *Im* represents the imaginary part of the term. The frequencies which can not be captured in the pyramid levels will be summarized in real valued high- and low-pass residuals $r_h$ and $r_l$, respectively. This decomposition of the image will be used as input to our network.

**Phase prediction.** The goal of our network is to predict the phase values of the intermediate frame, based on the steerable pyramid decomposition of the input frames. Each level of the multi-scale pyramid represents a band of spatial frequencies. The phase computation according to Equation (6) yields phase values between $[-\pi, \pi]$ for every pixel at each resolution.

We have seen earlier that there exists two solutions for the middle frame. Furthermore, the assumption that motion is encoded in the phase difference is only accurate for small motion, i.e. the lower levels of the pyramid. Due to the frequency banded filter design the response value is based on a locally limited spatial area. On the higher levels the motion could be larger than the receptive field of the filters. As a consequence, the phase values of a pixel at two different time steps are not comparable anymore. By assuming that large motion is already visible and captured correctly by the phase on a lower level, this information can be used to improve the prediction on the higher levels. Instead of using heuristics [29] to propagate the information upwards in the pyramid, we propose using a convolutional network to learn how to combine the available phase information.

## 4. Method

The aim of the network is to synthesize an intermediate image given its two neighboring images as input. Instead of directly predicting the color pixel values, our network predicts the values of the steerable pyramid decomposition.

### 4.1. Learning Phase-based Interpolation

The color input frames $I_1$ and $I_2$ are decomposed using the steerable pyramid (Eq. (2)). We denote the obtained decomposition as $R_1$ and $R_2$, respectively:

$$R_i = \Psi(I_i) = \{\{(\phi^i_{\omega,\theta}, A^i_{\omega,\theta})|\omega, \theta\}, r^i_l, r^i_h\} \ . \quad (7)$$

These decomposition responses $R_1$ and $R_2$ are the inputs to our network. Using these values, the objective is to predict $\hat{R}$, the decomposition of the interpolated frame. The prediction function, $\mathcal{F}$ is a CNN with parameters $\Lambda$. The interpolated frame $\hat{I}$ is given by

$$\hat{I} = \Psi^{-1}(\hat{R}) = \Psi^{-1}(\mathcal{F}(R_1, R_2; \Lambda)) \ , \quad (8)$$

where $\Psi^{-1}$ the reconstruction function.

The network is trained to minimize the objective function $\mathcal{L}$ over the dataset $\mathcal{D}$ consisting of triplets of input images $(I_1, I_2)$ and the corresponding ground truth interpolation frame, $I$:

$$\Lambda^* = \arg\min_{\Lambda} \mathbb{E}_{I_1, I_2, I \sim \mathcal{D}}[\mathcal{L}(\mathcal{F}(R_1, R_2; \Lambda), I)] \ . \quad (9)$$

Our objective is to predict response values $\hat{R}$ that lead to a reconstructed image similar to $I$. We also penalize the deviation from the ground truth decomposition $R$. This is reflected in our loss function that consists of two terms: an image loss and a phase loss.

**Image loss.** For the image loss we use the $\ell_1$-norm of pixel differences which has been shown to lead to sharper results than $\ell_2$ [25, 27, 32]:

$$\mathcal{L}_1 = ||I - \hat{I}||_1 \ . \quad (10)$$

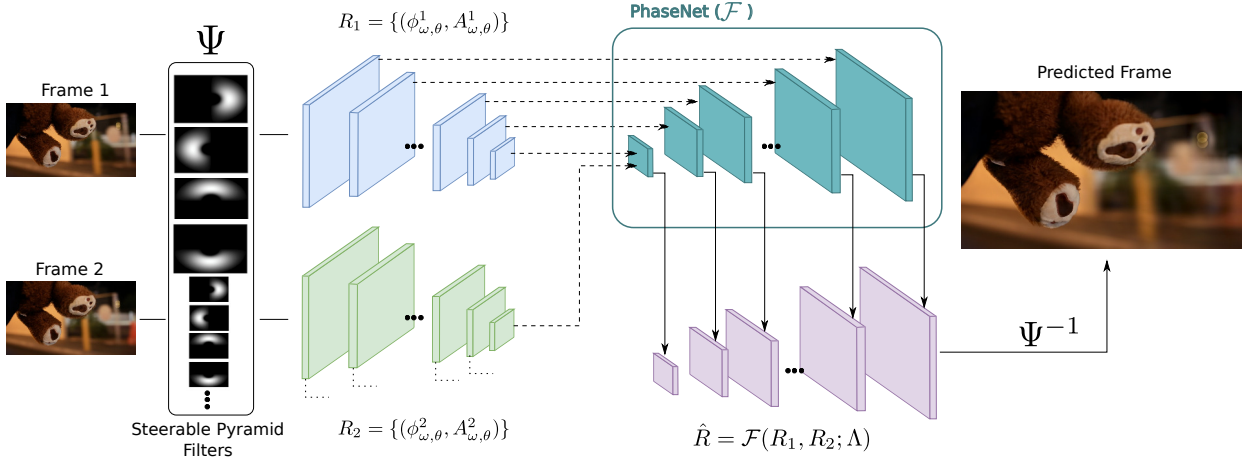

Figure 3: **PhaseNet architecture.** Given two consecutive frames, their decomposition can be obtained by applying the steerable pyramid filters ($\Psi$). The decomposition of these two input frames (denoted as $R_1$ and $R_2$) are the inputs to our network: PhaseNet, which has a decoder only architecture. The number of layers and their dimensions mirror the input frame decompositions. We only display the blocks of each level (the details of the blocks are discussed later). Each block takes as input the decomposition values from the corresponding level. We only display the links from the decomposition of the first frame to avoid cluttering the image. The predicted filter responses ($\hat{R}$) are then used to reconstruct the middle frame.

**Phase loss.** The predicted decomposition $\hat{R}$ of the interpolated frame consists of amplitude and phase values for each level and orientation present in the steerable pyramid decomposition. To improve the quality of the reconstructed images we add a loss term which captures the deviations $\Delta \phi$ of the predicted phase $\hat{\phi}$ from the ground truth phase $\phi$. The phase loss is then defined as the $\ell_1$ loss of the phase difference values over all levels ($\omega$) and orientations ($\theta$):

$$\mathcal{L}_{\text{phase}} = \sum_{\omega,\theta} ||\Delta \phi_{\omega,\theta}||_1 \;, \tag{11}$$

where $\Delta \phi$ is defined as

$$\Delta \phi = \text{atan2}(\sin(\phi - \hat{\phi}), \cos(\phi - \hat{\phi})) \;. \tag{12}$$

We use atan2, the four-quadrant inverse tangent, which returns the smaller angular difference between $\phi$ and $\hat{\phi}$.

We could also define a similar loss on the predicted amplitude values $\hat{A}_{\omega,\theta}$ but we found that it did not improve over the combination of phase and image loss in practice. As motion is primarily encoded in the phase shift, it is more important to enforce correct phase prediction.

We define our final loss as a weighted sum of the image loss and the phase loss:

$$\mathcal{L} = \mathcal{L}_1 + \nu \mathcal{L}_{\text{phase}} \;. \tag{13}$$

In our experiments the weighting factor $\nu$ is chosen such that the phase loss is one order of magnitude larger than $\mathcal{L}_1$, i.e. $\nu = 0.1$.

### 4.2. Network Architecture

The architecture of PhaseNet is visualized in Figure 3. The design is inspired by the steerable pyramid decomposition. For each resolution level it predicts the values of the corresponding level of the pyramid decomposition of the intermediate frame. It is structured as a decoder-only network increasing resolution level by level. At each level we incorporate the corresponding decomposition information from the input images. Besides the lowest level, due to the steerable pyramid decomposition, all other levels are structurally identical. At each level we also incorporate the information from the previous level. This follows the assumption that motion will be captured at different scales and the phase values do not differ arbitrarily from level to level.

As input to the network we use the response values from the steerable pyramid decomposition of the two input frames consisting of the phase $\phi_{\omega,\theta}$ and amplitude $A_{\omega,\theta}$ values for each pixel at each level $\omega$ and orientation $\theta$, as well as the low pass residual. Before passing them through the network we normalize the phase values by dividing by $\pi$. The residual and amplitude values are normalized by dividing by the maximum value of the corresponding level.

Each resolution level consist of a PhaseNet block (Figure 4) which takes as input the decomposition values from the input images, the resized feature maps from the previous level as well as the resized predicted values from the previous level. This information is passed through two convolution layers each followed by batch normalization [17] and ReLU nonlinearity [30], which have shown to help training. Each convolution layer produces 64 feature maps by either using $1 \times 1$ or $3 \times 3$ convolution filters (see supplementary material for details). In general, we observe, that smaller kernels are preferable for lower resolution. Between levels the resolution is increased by the scaling factor $\lambda$, which has been used to produce the steerable pyramid. Resizing is done by bilinear interpolation. On the lowest level, the first PhaseNet block receives as input only the concatenation of the two low level residuals of the two input frames.

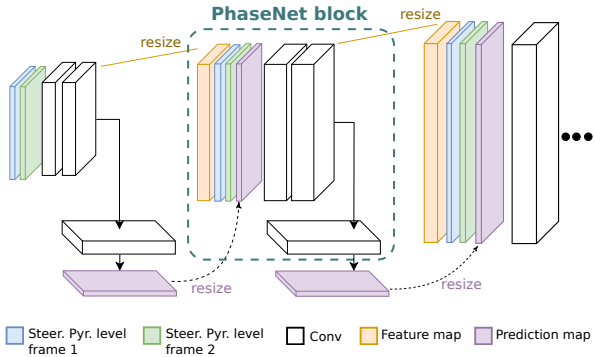

Figure 4: **PhaseNet block.** Each block of the PhaseNet takes as input the decompositions of the input frames at current level (shown in blue and green). Each level performs two successive convolutions with batch normalization and ReLU. From the intermediate features map, each block predicts the response (amplitude and phase) at current level with one convolution layer followed by the hyperbolic tangent function. Feature map and predicted values are reused in the next block after resizing.

After each PhaseNet block we predict the values of the in-between frame decomposition by passing the output feature maps of the PhaseNet block through one convolution layer with filter size $1 \times 1$ followed by the hyperbolic tangent function to predict output values within the range of $[-1, 1]$. From these we can compute the decomposition values $\hat{R}$ of the intermediate image and reconstruct it, see Section 4.3. The number of output channels depends on the number of predicted values for each pixel, i.e. $d$ for the lowest level, and $2bd$ for the intermediate levels, where we predict phase and amplitude for each dimension $d$ and orientation $b$.

In our case, the network is built for a single color dimension (i.e. $d = 1$) and trained for color images by reusing the weights across the color channels. This allows to significantly reduce the weights while producing comparable results. To process higher resolutions at testing time we share the weights of the highest three levels. We describe this in Section 5.

### 4.3. Image Reconstruction

In general we can reconstruct an image from the steerable pyramid decomposition by integrating over all pyramid levels according to Equation 1 and adding the low and high pass residual. Due to the normalization of the steerable pyramid values before passing them through PhaseNet and by predicting values between $[-1, 1]$ we need to remap the predicted values before we can reconstruct the image. The following remapping is applied to each pixel $(x, y)$ at each level $\omega$ and orientation $\theta$.

To compute the phase values $\hat{\phi}$ of $\hat{R}$ we scale the predicted values by multiplying them with $\pi$. To approximate the low level residuals and the amplitudes of the intermediate frame [29] propose to average the values. This works well for lower levels where these values correspond mainly to global luminance changes. For higher frequency bands, averaging the amplitude values can lead to artifacts. For more flexibility, instead of exactly averaging, we allow the network to learn the mixing factors.

The low level residual, $\hat{r}_l$ as well as the amplitude values $\hat{A}$ of $\hat{R}$ are computed using the predicted values as a linear scaling factor between the values of the input decompositions $R_1$ and $R_2$:

$$\hat{r}_l = \alpha * r_l^1 * (1 - \alpha) * r_l^2 , \qquad (14)$$
$$\hat{A} = \beta * A^1 + (1 - \beta) * A^2 , \qquad (15)$$

where $\alpha$ and $\beta$ are the learned mixing weights mapped to $[0, 1]$. We observe that the high pass residual can be ignored as the introduced blur is often very subtle.

### 4.4. Training and Implementation Details

Each pixel in the synthesized image is influenced by the predicted phase and amplitude values from all scales. For stability, we adopt a hierarchical training procedure where the layers at lowest levels are trained first. When training the first $m$ levels, we still need to reconstruct the interpolated frame to compute the loss. In this case we use ground truth response values for levels $m + 1, \ldots, n$ as illustrated in Figure 5.

This training procedure can be seen as a form of curriculum learning [7] that aims at improving training by gradually increasing the difficulty of the learning task. This type of learning strategy is often used in sequence prediction tasks and in sequential decision making problems where large speedups in training time and improvements in generalization performance can be obtained.

Our training procedure is related to the filtered scheme adopted in [13] where ground truth masks are first blurred then smoothly sharpened over time. In our case, by using a steerable pyramid decomposition we have already a coarse to fine representation of the image which is well suited for such a hierarchical training procedure. It also matches the assumption that the motion and therefore pyramid values of higher, finer levels are related to the previous, lower levels.

For training we use triplets of frames from the DAVIS video dataset [34, 33], randomly selecting patches of $256 \times$

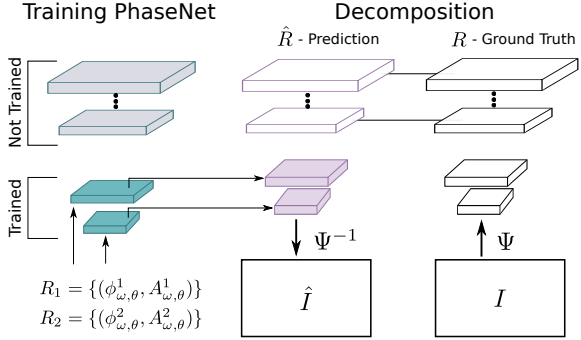

Figure 5: **Hierarchical training.** On the left, PhaseNet takes as input the decompositions $R_1$ and $R_2$ of the input frames. In this example the two lowest levels are being trained ($m = 2$). Corresponding blocks are displayed in green. The other blocks (in gray) will be added at the next iteration. On the right, we have the ground truth frame decomposition $R$. To reconstruct the predicted image, we use ground truth values for the layers not being trained yet.

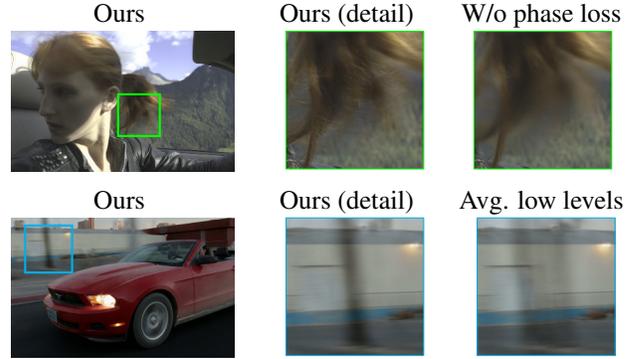

Figure 6: **Design choices.** The first row shows the benefit of using the phase loss giving sharper results compared to only using the image loss (best viewed on screen). For images larger than the training patches, the second row shows the benefit of reusing last layers weights over averaging the lowest levels of the decomposition. (Image source: [3, 22])

256 pixels. To build the pyramid decomposition we use a scale factor of $\lambda = \sqrt{2}$ leading to a pyramid of 10 levels. More details on the training procedure can be found in the supplementary material.

**Computation Time.** PhaseNet is implemented in Tensorflow and takes advantage of efficient spectral decomposition layers. With one Nvidia Titan X (Pascal), training our model (∼460k parameters) takes approximately 20h in total for 9 hierarchical training stages. Computation time for decomposition, interpolation and image reconstruction is $0.5s$ for $256 \times 256$ patches (training) and $1.5s$ for $2048 \times 1024$ images (testing).

## 5. Results

We compare our method with a representative selection of state-of-the-art methods by evaluating them quantitatively and qualitatively on various images. As a representative of optical flow we chose MDP-Flow2 [46], which currently performs best on the Middlebury benchmark for interpolation. To synthesize the interpolated frames from the computed optical flow field, we use the same algorithm as used in the benchmark [5]. According to Middlebury, MDP-Flow2 is followed closely by [32], a neural network based method learning seperable convolution filters for frame interpolation (SepConv). In terms of phase-based representation methods for frame interpolation we compare to [29] (Phase). The image sequences used are from the footage of [22], Blender Foundation [1], Vision Research [2] and YouTube [3, 4]. To produce the results of these methods, we use the code and trained models provided by the original authors.

**Loss function.** For training our network we use the combination of the two loss functions: the image loss ($\mathcal{L}_1$) and the phase loss ($\mathcal{L}_{phase}$). Training only with the image loss already produces reasonable interpolation results. Because the phase loss is computed at each resolution level and encodes motion relevant information, it is necessary to achieve sharp results, see Figure 6 (top). Furthermore, we observe that optimizing for the phase loss additionally to the image loss stabilizes the training procedure and helps to reduce training time. For our final results we use a linearly weighted combination of both terms, see Eq. (13). We did not notice any particular sensitivity of the results regarding the weighting factor ($\nu \in [0.1, 1]$). Using only the phase loss is however not sufficient.

**High resolution data.** Because we are using a fully convolutional network, we are able to handle larger images at testing time. Our network is trained on patches of $256 \times 256$ leading to a pyramid of 10 levels. To produce higher resolution images during testing, we need to extend the pyramid. We test our algorithm on images of $1280 \times 720$. For stability of the used Fast Fourier transform and the pyramid decomposition we symmetrically pad the images to $2048 \times 1024$ leading to 14 pyramid levels. A naive approach would be to consider averaging the phase values at the lower levels and use our model only on the 10 highest levels. However, this implicitly limits the range of motion we can interpolate, see Figure 6 (bottom right). A better approach is to reuse the weights of the trained higher levels for the following, additional layers. Because they shared their weights during training over several levels, this approach generalizes well to further levels, see Figure 6 (bottom middle).

**Qualitative comparisons.** We evaluate our method on a set of challenging image pairs including motion blur and extreme light changes, see Figure 10. Because optical flow based methods, such as MDP-Flow2, compute explicit pixel correspondences it produces visible artifacts once the used brightness constancy assumption is violated. The pure phase-based method as well our phase-based-network combined approach, on the other hand, are robust against such lighting changes and produce smooth and plausible results. In the case of the explosion scene in the second row, our result is even preferable over the pure phase-based approach.

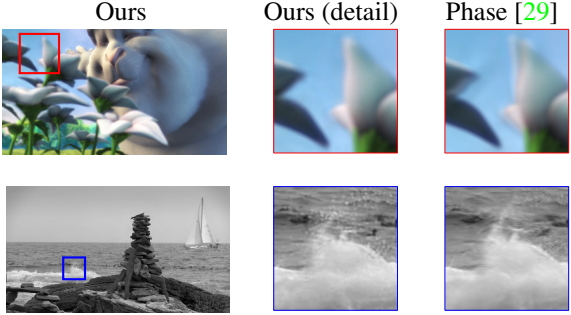

Figure 7: **Advantage of a data driven approach.** Using heuristics [29] for phase-based frame interpolation reaches its limits in these two examples. Our data driven approach is able to better handle large motion and obtains sharper results. (© Blender Foundation [1], © Vision Research [2])

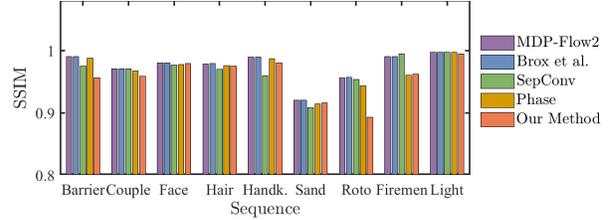

Figure 8: **Error measurements** of different methods for different sequences by computing the structural similarity measuremnt (SSIM) averaged over several frames. Example images of the evaluated sequences are shown in the supplementary material.

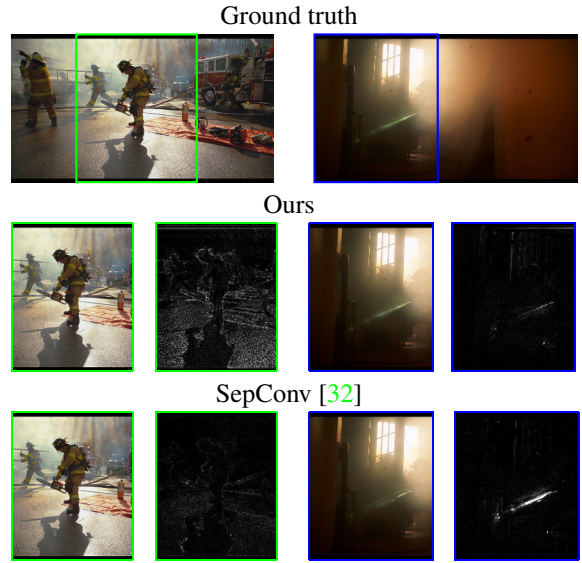

Figure 9: **Comparison** of interpolation results with our method and separable convolution filters to the ground truth including a difference map using absolute differences. Best viewed on screen. (© Vision Research [2])

The last two rows show some examples with motion blur. The pure phase-based approach is limited in the amount of motion it can handle. This is visible in the last row, where the pole in the background moves too far to be correctly captured by the method resulting in ghosting artifacts. In this example SepConv is unable to correctly interpolate the car due to the motion blur. Our method improves on both of them. However, the frequency banded filters influence some area around each in pixel in the spatial domain. As a result, reduced accuracy in the phase prediction can lead to some minor ringing and color artifacts during reconstruction. These are noticeable around high frequency edges. Although both phase-based methods have this issue in common, the main improvement of PhaseNet over the pure phase-based methods is visible in the case of interpolating large motion and high frequencies, as shown in Figure 7.

**Quantitative comparisons.** We use the same set of sequences as in [29], consisting of representative scenes with many moving parts and challenging lighting conditions as well as one synthetic example (*Roto*) containing many high frequencies. For quantitative evaluation, we compare several methods on a number of sequences using the leave-one-out method, where we compare synthesized frames to the original ones. In Figure 8 we report the error measurements using the structural similarity (SSIM) measure. In general, the optical flow method and SepConv achieve a better error measure, mainly due to the fact they introduce less blur. Especially for the sequences with high frequencies (*barrier*, *fireman*, *sand* and *roto*) we perform worse. The strength of our method lies in handling challenging scenarios with motion blur and brightness changes (e.g. *light* and *handkerchief*). Although the measure is perceptually motivated it does not always reflect the visual comparison, as illustrated in Figure 9. For the *light* sequence (right column), our approach produces noticeably better results. For the *fireman* sequence (left column), although the difference map shows a global degeneration for high frequency content for our method, there is no perceptual difference between the different methods.

**Discussion and limitations.** Our method significantly improves over previous phase-based methods, both in terms of motion range and high frequencies. It is well suited for scenes with motion blur and difficult light changes. We still however do not reach the same level of detail as methods which explicitly match and warp pixels. On the other hand these methods may produce more disturbing artifacts whereas our model creates less noticeable effects.

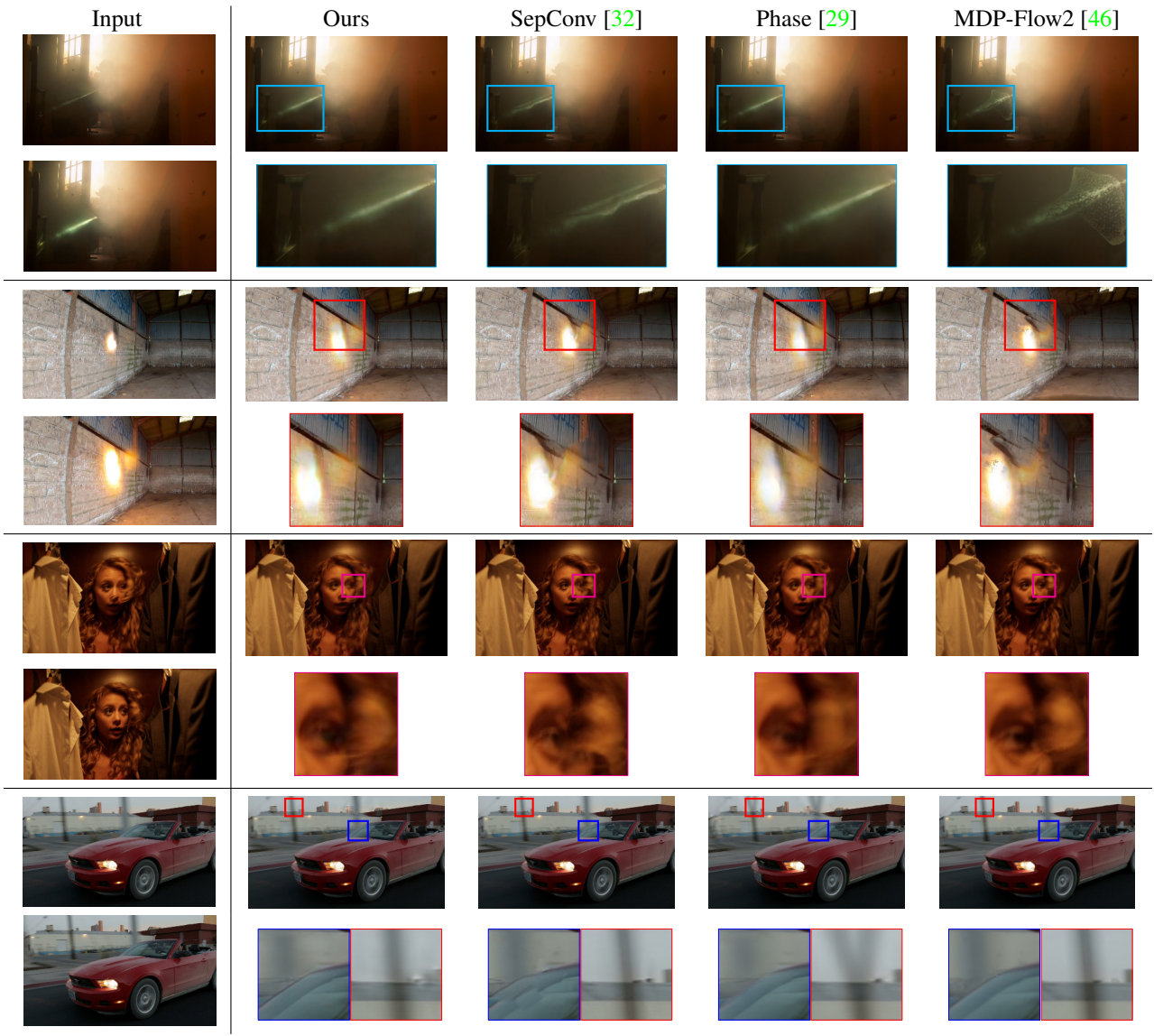

Figure 10: **Visual comparison** with frame interpolation methods on challenging scenarios. See text for details and discussion. (Image source: [2, 4, 22])

## 6. Conclusions

We have presented a method which combines the advantage of phase-based and data driven methods for frame interpolation. We propose a neural network architecture that synthesizes an interpolated frame from its predicted phase-based representation. By combining both a phase loss and standard $\ell_1$-norm over the reconstructed image we are able to produce visually preferable results over optical flow for challenging scenarios containing motion blur and brightness changes.

**Acknowledgments.** This work was supported by ETH Research Grant ETH-12 17-1.

## A. Error Measurements

In Table 1 we report the peak signal-to-noise ration (PSNR) in addition to the SSIM error reported in Figure 8. Example input images from the sequences used to compute these error measurements are shown in Figure 11.

## B. Details Network Architecture

The architecture of the PhaseNet consists of consecutive PhaseNet blocks. Figure 4 in the paper visualizes the concept of such a block. In Table 2 the specific details for each layer can be found. Each block consists of two convolution layers both followed by batch normalization and leaky ReLU nonlinearity with factor $0.2$. The prediction layers pred_$i$ consists of one convolution layer followed by the hyperbolic tangent function. pyr_$i$ summarizes the steerable pyramid decomposition information of the input images at the corresponding level $i$, i.e. low level residuals for $i = 0$ and phase and amplitude information for $i > 0$. To increase the resolution of the feature maps we use bilinear upscaling noted as *up()* in Table 2. Due to reusing the weights across the color channels and some of the layers, our network has only about $460k$ trainable parameters in total.

## C. Details Model Training

Our training dataset consists of about 10k triplets of frames from the DAVIS video dataset [34, 33]. At each iteration we randomly select patches of $256 \times 256$ pixels. We perform data augmentation through horizontal and vertical flipping of the patches.

We use Adam optimizer [19] with $\beta_1 = 0.9$, $\beta_2 = 0.999$ and learning rate $0.001$. The batch size used is 32 and reduced to 16 and 12, respectively, for the highest two training stages due to memory limitations. We train the lower levels for 12 epochs each and the highest two for 6 epochs due to the reduced batch size to have approximately the same number of iteration steps for each hierarchical level.

|  | Method | | | | |
| --- | --- | --- | --- | --- | --- |
|  | MDP-Flow2 | Brox *et al.* | SepConv | Phase | Ours |
| Barrier | 42.33 | 42.67 | 40.97 | 39.93 | 35.17 |
| Couple | 41.02 | 40.97 | 41.26 | 40.26 | 38.25 |
| Face | 40.94 | 40.89 | 40.60 | 40.28 | 40.31 |
| Hair | 37.15 | 37.69 | 37.00 | 36.86 | 36.34 |
| Handk. | 42.32 | 42.43 | 38.26 | 41.74 | 38.68 |
| Sand | 37.43 | 37.48 | 36.93 | 37.17 | 36.67 |
| Roto | 35.39 | 35.54 | 35.88 | 33.73 | 25.50 |
| Firemen | 40.26 | 40.21 | 42.54 | 33.80 | 34.78 |
| Light | 45.82 | 45.63 | 45.50 | 45.63 | 45.33 |

Table 1: **Error measurements** of different methods for the sequences shown in Figure 11 by computing the PSNR (higher is better).

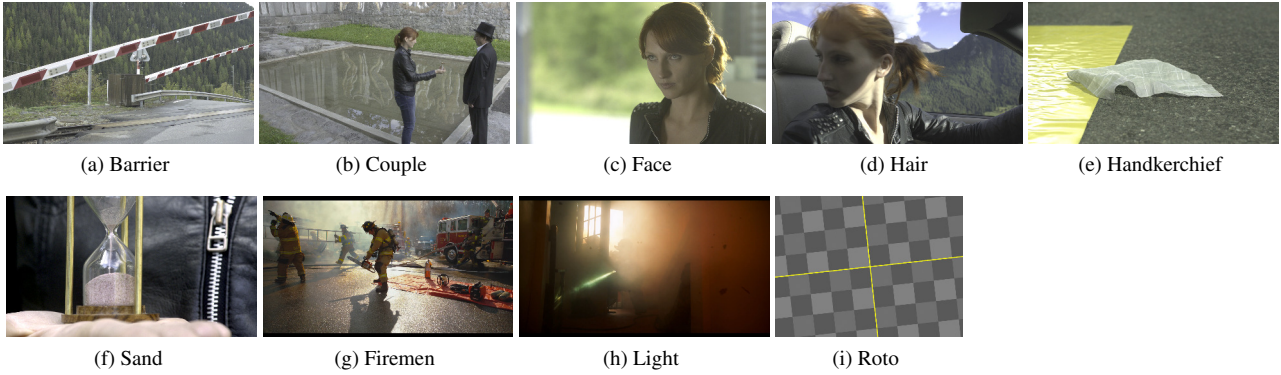

(a) Barrier  (b) Couple  (c) Face  (d) Hair  (e) Handkerchief

(f) Sand  (g) Firemen  (h) Light  (i) Roto

Figure 11: Example images from the sequences used for the error measurements.

| Name | Input | Kernel | Ch In/Out | Res | Reuse Weights |
|---|---|---|---|---|---|
| PhaseNetBlock_0 | pyr_0 | $1 \times 1$ | 2/64 | $8 \times 8$ | False |
| pred_0 | PhaseNetBlock_0 | $1 \times 1$ | 64/1 | $8 \times 8$ | False |
| PhaseNetBlock_1 | up(PhaseNetBlock_0)+up(pred_0)+pyr_1 | $1 \times 1$ | $(64+1+16)/64$ | $12 \times 12$ | False |
| pred_1 | PhaseNetBlock_1 | $1 \times 1$ | 64/8 | $12 \times 12$ | False |
| PhaseNetBlock_2 | up(PhaseNetBlock_1)+up(pred_1)+pyr_2 | $1 \times 1$ | $(64+8+16)/64$ | $16 \times 16$ | False |
| pred_2 | PhaseNetBlock_2 | $1 \times 1$ | 64/8 | $16 \times 16$ | False |
| PhaseNetBlock_3 | up(PhaseNetBlock_2)+up(pred_2)+pyr_3 | $3 \times 3$ | $(64+8+16)/64$ | $22 \times 22$ | False |
| pred_3 | PhaseNetBlock_3 | $1 \times 1$ | 64/8 | $22 \times 22$ | False |
| PhaseNetBlock_4 | up(PhaseNetBlock_3)+up(pred_3)+pyr_4 | $3 \times 3$ | $(64+8+16)/64$ | $32 \times 32$ | False |
| pred_4 | PhaseNetBlock_4 | $1 \times 1$ | 64/8 | $32 \times 32$ | False |
| PhaseNetBlock_5 | up(PhaseNetBlock_4)+up(pred_4)+pyr_5 | $3 \times 3$ | $(64+8+16)/64$ | $46 \times 46$ | False |
| pred_5 | PhaseNetBlock_5 | $1 \times 1$ | 64/8 | $46 \times 46$ | False |
| PhaseNetBlock_6 | up(PhaseNetBlock_5)+up(pred_5)+pyr_6 | $3 \times 3$ | $(64+8+16)/64$ | $64 \times 64$ | False |
| pred_6 | PhaseNetBlock_6 | $1 \times 1$ | 64/8 | $64 \times 64$ | False |
| PhaseNetBlock_7 | up(PhaseNetBlock_6)+up(pred_6)+pyr_7 | $3 \times 3$ | $(64+8+16)/64$ | $90 \times 90$ | False |
| pred_7 | PhaseNetBlock_7 | $1 \times 1$ | 64/8 | $90 \times 90$ | False |
| PhaseNetBlock_8 | up(PhaseNetBlock_7)+up(pred_7)+pyr_8 | $3 \times 3$ | $(64+8+16)/64$ | $128 \times 128$ | True |
| pred_8 | PhaseNetBlock_8 | $1 \times 1$ | 64/8 | $128 \times 128$ | True |
| PhaseNetBlock_9 | up(PhaseNetBlock_8)+up(pred_8)+pyr_9 | $3 \times 3$ | $(64+8+16)/64$ | $182 \times 182$ | True |
| pred_9 | PhaseNetBlock_9 | $1 \times 1$ | 64/8 | $182 \times 182$ | True |
| PhaseNetBlock_10 | up(PhaseNetBlock_9)+up(pred_9)+pyr_10 | $3 \times 3$ | $(64+8+16)/64$ | $256 \times 256$ | True |
| pred_10 | PhaseNetBlock_10 | $1 \times 1$ | 64/8 | $256 \times 256$ | True |

Table 2: **Details of the PhaseNet architecture.** The numbers of the channels and resolutions correspond to the case of using one color channel (weights reused for the other two) and a pyramid constructed with $\lambda = \sqrt{2}$ and 4 orientations. The + in the input column corresponds to concatenating the channels. In total the network has about 460k trainable parameters.